# MaskMol: Knowledge-guided Molecular Image Pre-Training Framework for Activity Cliffs with Pixel Masking

Zhixiang Cheng[1,2,3,#], Hongxin Xiang[1,3,#], Pengsen Ma[1], Li Zeng[3], Xin Jin[4], Xixi Yang[1], Jianxin Lin[1], Yang Deng[2], Bosheng Song[1], Xinxin Feng[2,5], Changhui Deng[2,*], Xiangxiang Zeng[1,*]

[1] College of Information Science and Engineering, Hunan University, Changsha, Hunan 410082, China

[2] Hunan Provincial Key Laboratory of Anti-Resistance Microbial Drugs, The Third Hospital of Changsha (the Affiliated Changsha Hospital of Hunan University), Changsha, Hunan 410015, China

[3] Department of AIDD, Shanghai Yuyao Biotechnology Co. Ltd, Shanghai, 201109, China.

[4] Ningbo Institute of Digital Twin, Eastern Institute of Technolog, Ningbo, 315200, China.

[5] State Key Laboratory of Chem-/Bio-Sensing and Chemometrics, Hunan Provincial Key Laboratory of Biomacromolecular Chemical Biology, and School of Chemistry and Chemical Engineering, Hunan University, Changsha, Hunan, 410082, China.

[#] These authors contributed equally.

*corresponding authors: xzeng@hnu.edu.cn, changhui-deng@cssdsyy.com

**Abstract**

Activity cliffs, which refer to pairs of molecules that are structurally similar but show significant differences in their potency, can lead to model representation collapse and make the model challenging to distinguish them. Our research indicates that as molecular similarity increases, graph-based methods struggle to capture these nuances, whereas image-based approaches effectively retain the distinctions. Thus, we developed MaskMol, a knowledge-guided molecular image self-supervised learning framework. MaskMol accurately learns the representation of molecular images by considering multiple levels of molecular knowledge, such as atoms, bonds, and substructures. By utilizing pixel masking tasks, MaskMol extracts fine-grained information from molecular images, overcoming the limitations of existing deep learning models in identifying subtle structural changes. Experimental results demonstrate MaskMol's high accuracy and transferability in activity cliff estimation and compound potency prediction across 20 different macromolecular targets, outperforming 25 state-of-the-art deep learning and machine learning approaches. Visualization analyses reveal MaskMol's high biological interpretability in identifying activity cliff-relevant molecular substructures. Notably, through MaskMol, we identified candidate EP4 inhibitors that could be used to treat tumors. This study not only raises awareness about activity cliffs but also introduces a novel method for molecular image representation learning and virtual screening, advancing drug discovery and providing new insights into structure-activity relationships (SAR).

# 1 Introduction

Drug discovery has always posed a significant challenge in life sciences, and its outcome could tremendously impact medical research. Recently, the advancements in machine learning and artificial intelligence are now opening up new possibilities and leading to breakthroughs in the field of drug discovery[1-3]. Over the past few years, machine learning has made remarkable advancements in various aspects of early drug discovery, such as molecular generation[4-6], molecular optimization[7-9], and molecular property prediction[10-16]. These technologies are offering more efficient and accurate methods for developing new drugs.

Molecular property prediction plays a vital role in the drug discovery and design process, as it directly impacts the safety, effectiveness, and efficiency of drug development[17]. The fundamental concept behind molecular property prediction is that molecules with similar structures tend to have similar properties[18]. As shown in **Figure 1a *Left***, molecules with distinct scaffolds exhibit different activities, and they can be well separated. However, there are cases called activity cliffs[19], where two molecules with similar structures have significantly different biological activities (**Figure 1b *Right***). Predicting activity cliffs holds substantial importance in rational drug design and the efficient discovery of new therapeutic agents[20]. Anticipating cliffs provides crucial insights into SAR and optimizes lead compounds more effectively, leading to more reliable biological activity prediction.

Activity cliff task is a very important yet understudied task in the field of drug discovery. Previous studies[21-23] have observed that graph-based models have poor performance on activity cliffs. We conjecture that the graph-based representation learning methods cannot separate two similar molecules in the feature space called representation collapse, resulting in poor performance on the activity cliffs. As shown in Figure b, we evaluate the performance of various GNN architectures on the activity cliffs, such as GCN[24], GAT[25], and MPNN[26]. The figure clearly shows that as the similarity between pairs of molecules becomes higher and higher, the distance in the feature space of graph-based methods decreases faster, which proves our conjecture. We defined this phenomenon as representation collapse. Therefore, we turn to discover other representations of molecules and find that various graph-based representation learning methods are inferior to image-based representation learning methods in identifying differences between similar molecules. Although molecular graphs and images describe the same molecular information, they are essentially different due to modal differences (graph versus image) and feature extraction differences (GNN versus CNN). In the activity cliff task, pairs of molecules have very similar structures and significant differences in activity. For example, a difference of just one atom can lead to a completely different activity. Increasing the discrimination between two similar molecules is the key to the success of predicting activity cliffs for deep learning models. For the GNN model, the small structural difference will be over-smoothed[27] out during information aggregation, resulting in little difference in the extracted features. This is also why GNN methods perform poorly on activity cliff tasks. For images, the convolution operation in CNN has the characteristics of local connectivity and parameter sharing, which makes the model pay more attention to local features to preserve these differences[28]. These observations indicate that image-based methods can amplify the differences between two similar molecules and motivate us to develop an image-based method for more accurate activity cliff prediction.

Besides, obtaining labels for activity cliffs requires expensive and time-consuming wet experiments. The inadequacy of labeled data significantly impacts model performance. Thus, we turn our attention to the pretrain-finetune paradigm[30-35], because the pre-training process doesn't need labels, and few labels can be used in the fine-tuning phase to enhance performance. However, unlike natural images, molecular images are not as information-dense and have many blank areas. If we simply apply the pre-trained framework in computer vision such as MAE[31] directly to molecular images, it would be challenging for the model to utilize meaningful molecular knowledge to identify subtle changes in cliff molecules. Therefore, it is necessary to use molecular domain knowledge to guide the model to learn molecule structures.

Moreover, activity cliffs often arise due to subtle changes at various molecular levels[36-37], such as specific atom substitutions, bond modifications, or functional group replacements. At the atomic level, substituting a hydrogen atom on a benzene ring with a chlorine atom can lead to significant changes in the molecule's binding interactions with receptors, thereby affecting its biological activity. Changing a single bond in a molecule to a double bond may alter the molecule's shape and electronic distribution, thereby affecting its interactions with targets and its biological activity. Replacing a hydroxyl group on a benzene ring with a methyl group. While the structural difference is insignificant, the hydroxyl group can form hydrogen bonds, significantly affecting the molecule's solubility and interactions with biological targets. As a result, our objective is to incorporate prior chemical knowledge into the model and utilize this activity cliff-related knowledge to instruct the model in learning molecules. Here, we present a novel self-supervised pre-training framework called MaskMol, which focuses on learning fine-grained representations from molecular images with knowledge-guided pixel masking. We design three pixel masking-based pre-training tasks with three different levels of knowledge, involving atomic knowledge, bond knowledge, and motif knowledge. These tasks enable MaskMol to comprehensively learn the local regions of molecules by pixel-level knowledge prompts.

In summary, our main contributions are:

- We first pinpoint the bottleneck in the molecular activity cliff task that the cliff molecules give rise to deep learning model representation collapse. Image-based model is superior to graph-based model due to alleviating representation collapse.

- We design a novel and multi-level knowledge-guided molecular image self-supervised learning framework (called MaskMol) using a pixel masking strategy. After pre-training on a large-scale dataset consisting of approximately two million molecules, MaskMol demonstrated a significant performance enhancement on activity cliff estimation datasets and compound potency prediction datasets.

- Explainable case study and visualization demonstrate that MaskMol strongly enables cliff awareness for bioactivity estimation and extracting meaningful SAR information for intuitive interpretation.

- Through MaskMol, we identified candidate EP4 inhibitors that could be used to treat tumors, demonstrating that MaskMol can be used as a promising method under activity cliff virtual screening scenario.

## 2 Results and Discussion

### 2.1 Overview of MaskMol

This section gives an overview of our MaskMol, highlighted in **Figure 1c** and **Figure 1d**. To accurately estimate molecular activity cliffs, we developed a knowledge-guided molecular image pre-training framework by fine-grained pixel masking, MaskMol. It consists of two parts: (1) three knowledge-guided pixel masking strategies, and (2) three knowledge-guided masked pixel prediction tasks for pre-training. See the Experimental Section for more descriptions on MaskMol.

Firstly, the conversion from molecular SMILES to molecular images is performed using RDKit. To eliminate any extraneous color effects, we proceed by removing all non-essential hues from the molecular images. Next, again leveraging RDKit, we apply green hues to atoms, bonds, and motifs separately. In the following, HSV detection isolates regions with green pixels within the highlighted image. To introduce an element of randomness, we select a subset of atom/bond masking images by randomly choosing a fraction (determined by the masking ratio $\gamma$) from the available masking atom/bond image sets.

Consequently, we generate a set of masking images, totaling $\gamma \cdot N_{atom}$ and $\gamma \cdot N_{bond}$ in number. It is important to note that to ensure that motifs do not cross each other, we only randomly select one masking image from the set of masking motif images. Moving forward, the masked image is combined with the original molecular image. Precisely, we adjust the white region of the masking image to correspondingly modify the region of the molecular image. In this synthesis process, we end up with three masked molecular images: the masked atom/bond/motif image. The three images are input through ViT to obtain latent features and classified through different fully connected layers. The pre-trained molecular encoder is fine-tuned on downstream tasks to further improve model performance.

## 2.2 Model Performance on Downstream Tasks

To evaluate the effectiveness of the image-based representations learned by MaskMol, we choose wide-ranging popular or state-of-the-art baselines for comparison on activity cliff estimation benchmark (ACE) called MoleculeACE[21], including 12 pre-training baselines and 11 traditional machine learning methods.

We refer to the original paper[21] and follow its strategy to split the dataset. To assess the generalization of MaskMol, we employed a widely used splitting strategy known as scaffold split on ACE task and compound potency prediction (CPP) task. This is a more challenging but practical setting since the test molecules can be structurally different from the training set.

### 2.2.1 Activity Cliff Estimation

As shown in **Figure 2a**, we compared the performance of MaskMol with three types of state-of-the-art self-supervised molecular representation models: (1) sequence-based, (2) graph-based, and (3) image-based models. MaskMol has a better performance compared with sequence-based (for example, ChemBERTa[38]), graph-based (for example, GROVER[39], MolCLR[40], GEM[41], EdgePred[42], Mole-BERT[10], 3DInformax[43], GraphMVP[44], and InstructBio[45]), and image-based models (for example, ImgaeMol[17]) using MoleculeACE experimental set-up. Compared with the second-best model (InstructBio), the elevated RMSE of MaskMol ranges from 2.3% to 22.4% with an overall relative improvement of 11.4% across 10 ACE datasets, in particular for HRH3 dataset (19.4%

RMSE improvement) and ABL11 dataset (22.4% RMSE improvement). In addition, MaskMol achieved lower RMSE values (**Figure 2b**) on D4R (RMSE = 0.73), DAT (RMSE = 0.59), FX (RMSE = 0.73), GSK3 (RMSE = 0.69), HRH3 (RMSE = 0.58), SOR (RMSE = 0.76), ABL11 (RMSE = 0.66), GR (RMSE = 0.68), CLK4 (RMSE = 0.85), and OX2R (RMSE = 0.67) compared with traditional ECFP-based methods across multiple machine learning algorithms, including support vector machine[46], random forest[47], k-nearest neighbors[48], multilayer perception[49], and gradient boosting machine[50]. In summary, Our method MaskMol, surpasses other state-of-the-art methods, achieving the lowest RMSE in these comparisons. To further substantiate MaskMol's efficacy in identifying activity cliff pairs, we showcase results using $RMSE_{Cliff}$ as an additional performance metric. On the DAT and OX2R datasets, MaskMol achieves a 6.7% improvement in $RMSE_{Cliff}$ compared to the second-best method ($SVM_{ECFP}$). Taking into account the two metrics of RMSE and $RMSE_{Cliff}$, MaskMol also has a lower value than any other state-of-the-art molecular representation models (**Figure 2c**). Furthermore, to evaluate the disparity between the prediction and label, we employ Kullback-Leibler Divergence (KLD[51]) for measuring distribution differences (**Figure 2d**). The KLD values of all ACE datasets are significantly lower and the distributions of label and prediction values are close, except CLK4. We hypothesize that the relatively pronounced discrepancies observed in the CLK4 dataset could be attributed to its limited molecule number (731), which may have resulted in an under-fitted model.

To test the generalization of MaskMol, we split the datasets using a scaffold split (**Figure 3a**). We found that MaskMol significantly outperforms $SVM_{ECFP}$ models across all 10 ACE datasets. For instance, the RMSE values of MaskMol (RMSE = 0.69) compared with $SVM_{ECFP}$ model (RMSE = 0.97) in the prediction of ABL11 are elevated by over 28.9%. We further evaluated the $RMSE_{Cliff}$, compared with $SVM_{ECFP}$ models, MaskMol achieves better performance with a performance advantage of 6.4% on average, in particular for SOR (20.9% $RMSE_{Cliff}$ improvement). Compared with the molecule image pre-training model (ImageMol), the elevated RMSE of MaskMol ranges from 6% to 28.8% with a performance advantage of 17% on average, the elevated $RMSE_{Cliff}$ ranges from 9.4% to 40% with a performance advantage of 19.4% on average.

These results validate MaskMol's ability to precisely predict molecules exhibiting activity cliffs. Notably, ECFP-based methods demonstrate robust performance, whereas graph-based methods tend to underperform in activity cliff estimation. Graph-based models are vulnerable to representation collapse when faced with activity cliffs, and they face challenges in learning from non-smooth objective functions[23]. Furthermore, we found that image-based methods such as ImageMol have lower RMSE and $RMSE_{Cliff}$ than graph-based algorithms (EdgePred, GraphMVP, 3DInfomax, Mole-BERT). This further demonstrates that the CNN-based model can use local inductive biases to identify subtle cliff changes. Although InstructBio attempts to mitigate representation collapse by leveraging a substantial amount of unlabeled data as pseudo-labels, it still does not match the performance of ECFP-based methods. The addition of pseudo-labels helps to clarify class boundaries[52], suggesting that semi-supervised learning could emerge as a novel solution for addressing activity cliffs.

**2.2.2 Compound Potency Prediction**

Although MaskMol is primarily designed for solving fine-grained tasks such as ACE, it also performs well on the coarse-grained task of CPP. Compound potency prediction is crucial to the drug discovery and design process[53-54]. Researchers aim to forecast the biological activity of chemical compounds, explicitly measuring their potency in terms of the amount needed to produce a desired effect. As shown in **Figure 3b**, MaskMol has a better performance compared with sequence-based (ChemBERTa), graph-based (MolCLR, MGSSL, MPG[55], and GraphMVP), and image-based models (ImageMol) using a scaffold split. Notably, on the BACE1 dataset, MaskMol achieves a small MAE of 0.56, while the best-performing baseline model (ImageMol), achieves 0.63. It is worth mentioning that MaskMol achieves this performance using only 2M pre-training data, compared to the 10M pre-training data used by ChemBERTa and ImageMol. This demonstrates that MaskMol can achieve superior performance with significantly less pre-training data.

**2.3 Ablation Studies on MaskMol**

We perform comprehensive experiments to investigate the impact of each component in MaskMol on the activity cliff estimation. As illustrated in **Figure 3c**, seven out of ten datasets have a pre-training gain of more than 30% and the gain reaches its peak at 45.87% on the DAT dataset. Furthermore, the average gain across all ACE datasets surpasses 34.43%, underscoring the substantial enhancement in MaskMol's performance attributable to knowledge-guided masked pixel prediction tasks. Unlike graphs, graph treats molecules as nodes and bonds, which encode a large amount of chemical information such as atom types and bond types. For an initialized image model, molecules are input into the model in the form of RGB pixels, which do not contain any chemical information. The model's understanding of molecular images is limited to the fact that the image is composed of some "line." Therefore, it is necessary to help the model understand the chemical information in the image, which allows the model to understand the specific meaning of the "lines." in the image. This is why we can see that MaskMol has greater improvement gains than graph-based GROVER before and after pre-training (34.43% versus 8.53%). The observed decline in performance for "w/o AMPP" (RMSE 4.5% decline), "w/o BMPP" (RMSE 16.4% decline), and "w/o MMPP" (RMSE 21% decline) indicates that the removal of any level knowledge-guide task adversely affects MaskMol's performance, with MMPP being the most influential. We also explored the impact of pre-training with different data scales. The size of pre-training dataset for $MaskMol_{base}$ and $MaskMol_{small}$ are 20K and 2M respectively. We found that the average RMSE performance increased from 0.76 to 0.70 as the pre-trained data scale increased. This suggests that MaskMol will be further improved as more molecules are added to the pre-training dataset.

Additionally, we delve into analyzing the implications of the masking ratio, examining how its value affects MaskMol's overall performance (**Figure 3d**). It is worth noting that the optimal masking ratio in our study significantly deviates from the typical ratios used in BERT and MAE. BERT typically employs a masking ratio of 15%, whereas MAE utilizes a masking ratio as high as 75%. However, we found that a 50% masking ratio yields optimal results in our experiments. Molecular images are rather sparse with most pixels being empty and the resolution of the images is important in such settings. Thus, we research the impact of image size and the ratio of empty spaces to useful pixels on the learned representations (**Figure 3e**). The results show that the image size and useful pixel ratio achieved similar performance on ACE dataset (p>0.05, Mann-Whitney U test[56]).

**2.4 Interpretation of MaskMol**

**2.4.1 Investigation of MaskMol Representation**

We use t-SNE to compare the representation learned by MaskMol with the ECFP fingerprints feature (**Figure 4a,b**). The t-SNE algorithm maps similar molecular representations to adjacent points in two dimensions. We observe that ECFP can only be mapped based on structure, resulting in active and inactive molecules being mixed in the feature space. Through multi-level knowledge-guided masked pixel prediction tasks, MaskMol can be aware of changes in atom/bond/motif when any atom/bond/motif in the image changes. Thus, the representations learned by MaskMol can effectively distinguish between active and inactive molecules, with a clear boundary between them.

Additionally, we have included some randomly selected pairs of activity cliffs in the figure to illustrate the similar and dissimilar molecules learned by MaskMol based on their biological activity. MaskMol can learn similar representations from molecules with similar structures and properties and map molecules with significant differences in structures and properties to distinct feature spaces. This demonstrates that MaskMol learns the topological structure information between molecules and uses properties to differentiate between molecules.

To measure the distance between active cliff pairs in feature space, we introduce a distance metric $d = \frac{1}{N}\sum_{i=1}^{N}\rho_i$, $\rho_i = \sqrt{(x_1 - x_2)^2 + (y_1 - y_2)^2}$, where $M_1 = (x_1, y_1), M_2 = (x_2, y_2)$ are active cliff pair coordinates in the unified feature space. **Figure 4c** illustrates that in all ACE datasets, the distance between active cliff pairs in the feature space generated by MaskMol is considerable, significantly greater than that of ECFP. This observation highlights the effectiveness of MaskMol in accurately estimating activity cliffs, as it can capture subtle structural variations and utilize them to describe and represent molecules.

**2.4.2 Explaining MaskMol via Attention Visualization**

We applied three levels of knowledge-guided pixel masking to the molecular images and used Grad-CAM[57] to visualize the areas of attention (**Figure 4d**). The results show that MaskMol accurately classifies the knowledge and focuses on the appropriate masked areas. This indicates that our three knowledge-guided masked pixel prediction tasks allow the model to identify different molecular chemical structures.

In **Figure 4e**, we provide a comparative analysis of key substructures associated with activity cliffs, as extracted by various deep learning (DL) methods. We select the top-3 most crucial edges detected by PGExplainer[58]. GNNs tend to allocate attention to insignificant regions of the cliff molecule and emphasize the identical structure. This observation supports our hypothesis that GNNs are susceptible to representation collapse when dealing with active cliffs, thereby hindering their ability to correctly identify cliff molecules. We can see that ImageMol focuses on large areas of the molecules, while MaskMol, without pre-training, only focuses on the entire molecule and ignores irrelevant blank areas. However, neither of them pays attention to the important substructure that affects the activity. MaskMol successfully identifies the most informative substructure and judges compound activity based on these substructures. These plots convincingly prove that MaskMol recognizes subtle differences in activity cliff pairs' substructures and can provide reliable and informative insights for medicinal chemists in identifying key substructures.

**2.4.3 Explaining MaskMol via Attention Visualization**

We use Substructure-Mask Interpretation (SME[59]) to further quantify the contribution of substructure to MaskMol predictions. We define the impact of the masking substructure on the overall prediction as the attribution. We make two predictions with MaskMol, one before and one after applying the substructure masking to the molecular image, and consider the difference between the predicted values as the attribution $\text{Attribution}_{sub} = f(x) - f(x_{sub})$, among them, $x$ represents the molecular image, $x_{sub}$ represents the molecular image of masking substructure, and $f$ represents MaskMol. By calculating the contribution of substructure to model predictions, we can gain insight into the impact of substructure on molecule activity. As depicted in **Figure 5a**, adding substructures such as benzene ring (Attribution = -1.93, $K_i$ = 5,370 nM) and ethyl alcohol (Attribution = -0.95, $K_i$ = 758 nM), the attributions are lower than zero, and the influence of the benzene ring is greater than that of ethyl alcohol, which is highly consistent with the molecular activity value. It can also be found that the position of the propyl group affects the activity, and the attribution value also makes the same judgment. **Figure 5b** also shows the same conclusion in the DAT dataset. In addition to biological activity, we also present the chemically intuitive explanation of MaskMol on Mutagenicity. **Figure 5c** and **Figure 5d** display the analysis of different substructures based on their Mutagenicity. A positive attribution indicates that the substructure contributes to toxicity, while a negative attribution suggests that the substructure has a detoxifying effect. **Figure 5c** reveals that nitro, amino, and quinone groups enhance the model's ability to predict toxicity, while carboxyl groups improve the model's prediction of non-toxicity. This observation aligns with previous studies, which have identified aromatic nitro, aromatic amino, and quinone groups as toxic and carboxyl groups as detoxifying[60-62].

In summary, this visualization provides evidence that MaskMol is subtle structure-aware and exploits structural differences to make accurate predictions. Thus, MaskMol can provide meaningful and fresh SAR insights to help medicinal chemists in structural optimization and de novo design.

**2.5 Virtual screening using MaskMol**

EP4 receptor has been widely investigated and recognized as a promising drug target for cancer immunotherapy[63]. We manually collected data from multiple sources, including the BindingDB[64], ChEMBL database, and patent libraries targeting EP4. Canonicalization of the molecules was achieved utilizing RDKit, and duplication of SMILES was deleted, resulting in a finalized dataset comprising 1633 molecules. We evaluated the performance of MaskMol on EP4 targets with a random split of 8:1:1. We found that MaskMol has a low RMSE on the test set (RMSE = 0.577), and the prediction values are linearly correlated with the label values in **Figure 5f *Left*** ($R^2$ = 0.789). The t-SNE visualization in the latent space showed a clear boundary between inhibitors and non-inhibitors (**Figure 5e** grey dots). To test the generalization ability of MaskMol, we constructed an additional patent set (131 molecules) from the extended patents and literature as an external validation set ($R^2$ = 0.755). We found that inhibitors and non-inhibitors in the patent test were also perfectly separated. MaskMol identified 9 known EP4 inhibitors and visualized these 9 molecules to embedding space (**Figure 5e**), suggesting structural identification ability of MaskMol to learn discriminative information. These nine molecules (Grapiprant[65], L001[66], CJ-042794[67], MK-2894[68], CR6086[69], ONO-4578[70], E7046[71], HL-43[72], and AMX12006[73]) have been validated (including cell assay, clinical trial, or other evidence) as potential EP4 inhibitors. These findings demonstrate the ability of MaskMol to provide robust and generalizable molecular representation and prediction of inhibitors of targets, making it an efficient and effective virtual screening method.

## 3 Discussion

In the field of early-stage drug discovery, machine learning is gaining prominence, yet the concept of activity cliffs remains underexplored. Activity cliffs, which refer to structurally similar molecules with significant differences in potency, are critical for virtual screening and developing models that understand complex structure-activity relationships. Traditional graph-based methods often struggle with representation collapse due to high similarity between activity cliffs. To address this, we developed MaskMol, a knowledge-guided self-supervised learning framework utilizing molecular images. MaskMol employs three pre-training tasks with pixel masking, incorporating atomic, bond, and motif knowledge. This approach enables MaskMol to effectively learn local molecular regions and detect subtle changes in activity cliffs. Experimental results confirm MaskMol's superior accuracy in predicting activity cliffs and its performance compared to other state-of-the-art algorithms. Extensive experiments and ablation studies validate the effectiveness of each MaskMol component and determine the optimal ratio for knowledge-guided pixel masking. Furthermore, MaskMol identifies critical substructures responsible for activity cliffs through visualization, enhancing researchers' understanding of compounds and facilitating the drug discovery process. This study not only raises awareness about activity cliffs but also introduces a novel method for molecular image representation learning and virtual screening, advancing drug discovery and providing new insights into structure-activity relationships.

Future potential directions may improve MaskMol further: (1) Incorporating more chemical knowledge (such as fingerprints knowledge, 3D space structure knowledge, and chemical reaction knowledge) into image model is a promising future direction. Fingerprint-based methods have demonstrated excellent performance in predicting activity cliffs. Therefore, incorporating information from multiple fingerprints, such as MACCS[74], ECFP, PharmPrint[75], and USRCAT[76], into the image model can enhance its accuracy. Currently, the consideration of activity cliffs does not include chiral cliffs. Learning 3D spatial structure information may be beneficial for predicting chiral cliffs. Despite the structural similarity of cliff molecules, their reaction synthesis processes differ, providing a unique perspective that can be used to identify these molecules. (2) Fine-grained alignment of images and other representations (for example SMILES and graph). By applying our proposed knowledge-guided pixel masks, we can perform fine-grained masking on images. This approach allows us to align images and graphs (or SMILES) more precisely, guiding the model to learn from multiple views and improving its ability to capture subtle differences. (3) Multi-task learning of multiple activity cliff prediction datasets. This approach will not only predict activity cliffs but also address related tasks such as toxicity prediction and pharmacokinetic parameter prediction. By tackling multiple tasks simultaneously, we expect to improve the model's overall generalization and performance.

## 4 Methods

### 4.1 Data and code availability

All of the codes are freely available at GitHub: https://github.com/ZhixiangCheng/MaskMol. The datasets for activity cliff estimation can be downloaded from MoleculeACE at the following URL: https://github.com/molML/MoleculeACE/tree/main/MoleculeACE/Data/benchmark\data.

The datasets for compound potency prediction can be obtained at the following URL: https://github.com/TiagoJanela/ML-for-compound-potency-prediction/tree/main/dataset.

The Mutagenicity datasets used in this study are available at https://doi.org/10.5281/zenodo.7707093.

### 4.2 Knowledge-guided Masked Pixel Prediction

**Definition.** A molecule's 2D information can usually be represented as a graph $G = (\mathcal{V}, \mathcal{E})$ with atoms $\mathcal{V}$ as nodes and the edges $\mathcal{E}$ given by covalent bonds. But in our experiments, the molecule is expressed as the image $x \in \mathbb{R}^{H \times W \times C}$, where $(H, W)$ is the resolution of the molecular image, C is the number of channels.

**4.2.1 Atom-level Masked Pixel Prediction**

We counted the atom types of molecules in the pre-training data and selected the ten most frequent atom types (e.g., C, N, O, Cl). Correspondingly, the ten atom types serve as pseudo-labels for the atom-level masked pixel prediction (AMPP). Formally, the molecular image set and the pseudo-labels are $\{x_i \in R^{224 \times 224 \times 3}\}_{i=1}^{N}$ and $y^{atom} \in \{0,1,\cdots,9\}^{10}$ respectively. For each $x_i$, we will get the mask atom image sets $M = \{M_j\}_{j=1}^{N_{atom}}$ by Masking. Random sampling $M$ with a masking ratio $\gamma$ to get the subset of $M$ as $M^* = \{M_j\}_{j=1}^{m}$, where $m = \gamma \cdot N_{atom}$ denotes the masking image number of subset. Then, we can obtain the masking atom image, denoted as $\tilde{x}_i = x_i \ominus M^*$, where $\ominus$ indicates modifying the pixel value in $x_i$ corresponding to the white pixel area in *M* to white. Following ViT[32], we divide a masking atom image $\tilde{x}_i$ into regular non-overlapping patches. To save calculation time and make our model pay more attention to the masked patches, we only calculate the loss of the masked patches $\Omega(\tilde{x}_i)$. Finally, the cost function of the AMPP task is as follows:

$$\mathcal{L}_{AMPP} = \arg\min_{\theta, W} \frac{1}{N} \sum_{i=1}^{N} \ell(\omega(f_\theta(\Omega(\tilde{x}_i))), y^{atom})$$

where $f_\theta$ and $\theta$ refer to the mapping function and corresponding parameters of the molecular encoder, $\omega$ represents the parameters of the fully connected classification layers, $\ell$ is the cross-entropy (CE) loss function.

**4.2.2 Bond-level Masked Pixel Prediction**

The workflow of the bond-level masked pixel prediction (BMPP) is similar to that of AMPP, and the difference is that there are only four bond types, i.e., single, double, triple, and aromatic, and the pseudo-labels are $y^{bond} \in \{0,1,2,3\}^4$. The BMPP loss function is defined

as follows:

$$\mathcal{L}_{\text{BMPP}} = \arg\min_{\theta,W} \frac{1}{N}\sum_{i=1}^{N} \ell(\omega(f_\theta(\Omega(\tilde{x}_i))), y^{bond})$$

**4.2.3 Motif-level Masked Pixel Prediction**

Breaking of Retrosynthetically Interesting Chemical Substructures (BRICS[77]) based on chemical reaction templates was utilized to partition functional groups. However, the functional group vocabulary obtained through the BRICS division is somewhat redundant. To address this issue, two rules defined in MGSSL[12] were applied to eliminate redundant functional groups. As a result, we obtained a motif vocabulary consisting of 9854 motifs. We opted for the top 200 motifs with the highest occurrence and eliminated molecules lacking these particular motifs to reduce time and space burdens on the MMPP task. The motif-level masked pixel prediction (MMPP) process is also consistent with AMPP. The difference is that the pseudo-labels are $y^{motif} \in {0,1,\cdots,199}^{200}$ and we only randomly sample a motif in $M$ as $M^*$ so that there is no intersection between motifs and the model can extract accurate motif information. It is worth noting that when calculating loss, we use the classification token feature $\tilde{x}_i^{cls}$ to classify and perform loss calculation. The MMPP loss function is defined as follows:

$$\mathcal{L}_{\text{MMPP}} = \arg\min_{\theta,W} \frac{1}{N}\sum_{i=1}^{N} \ell\left(\omega\left(f_\theta\left(\tilde{x}_i^{cls}\right)\right), y^{motif}\right)$$

**4.3 Pre-training and Fine-tuning**

Here, we used ViT as our molecular encoder. After using data augmentations and masking to obtain masking molecular images $\tilde{x}_i$, we forward these images $\tilde{x}_i$ to the ViT model to extract latent features $f_\theta(\tilde{x}_i)$. Then, these latent features are used by three pretext tasks to calculate the total cost function $\mathcal{L}$, which is defined as

$$\mathcal{L} = \mathcal{L}_{\text{AMPP}} + \mathcal{L}_{\text{BMPP}} + \mathcal{L}_{\text{MMPP}}$$

In order to pretrain our MaskMol, we first gathered 2 million unlabeled molecules with drug-like properties from the PubChem database[78]. We divided the 2M pre-training data into a training set (95%) and a validation set (5%), and judged the pre-training performance through the accuracy of each task. Finally, the AMPP, BMPP, and MMPP accuracy can reach 99.3%, 98.0%, and 89.6%, respectively. After the initial pre-training phase, we proceed to fine-tune the pre-trained encoder for the specific downstream tasks. In particular, we incorporate an extra fully connected layer after the encoder. The output dimension of this layer is set to match the number of categories associated with the downstream tasks.

### 4.4 Training Details

#### 4.4.1 Baselines

The performance regarding methods (MLP[49], GBM[50], RF[47], SVM[46], KNN[48], AFP[79], MPNN[26], GAT[25], GCN[24], CNN[80], LSTM[81] is derived from MoleculeACE[21]. The performance regarding methods (MolCLR[40], GROVER[39], GEM[41], InstructBio[45]) is derived from InstructBio. We additionally execute experiments on the activity cliff estimation datasets following the same experimental setting used in Mole-BERT[10], EdgePred[42], GraphMVP[44], 3DInfomax[43], and ImageMol[17].

#### 4.4.2 Evaluation Metrics

The overall performance of MaskMol was quantified via the mean absolute error (MAE) or root-mean-square error (RMSE) computed on the bioactivity values (i.e., $pK_i$ or $pIC_{50}$):

$MAE = \frac{1}{n}\sum_{i=1}^{n}|y_i - \hat{y}_i|$, $RMSE = \sqrt{\frac{1}{n}\sum_{i=1}^{n}(\hat{y}_i - y_i)^2}$, where $\hat{y}_i$ is the predicted bioactivity of the i-th molecule, $y_i$ is the corresponding experimental value, and n represents the total number of molecules. On activity cliffs, the performance of MaskMol was quantified by computing the root-mean-square error ($RMSE_{cliff}$) on compounds that belonged to at least one activity cliff pair: $RMSE_{cliff} = \sqrt{\frac{1}{n_c}\sum_{i=1}^{n_c}(\hat{y}_i - y_i)^2}$, where $\hat{y}_i$ is the predicted bioactivity of the $i$-th compound, $y_i$ is the corresponding experimental value, and $n_c$ represents the total number of compounds on activity cliffs.

**4.4.2 Hyperparameter**

MaskMol is pre-trained by SGD optimizer with a learning rate of 0.01, weight decay 10-5, momentum 0.9, and batch size 128 for approximately 2 days with 4 NVIDIA A100 GPU (40GB). In downstream tasks, the pre-trained model is fine-tuned using SGD optimizer with batch size [8, 16, 32, 64], learning rate [5e-5, 5e-4, 5e-3], weight decay 10-5, momentum 0.9 on Ubuntu 18.04.1 with 15 vCPU Intel(R) Xeon(R) Platinum 8375C CPU @ 2.90GHz and NVIDIA 4090 (20GB).

**4.5 Activity Cliff Estimation Dataset**

We adopt the MoleculeACE[21] framework, which focuses on identifying activity cliffs and quantifying model efficacy. Pairs of molecules that exhibit computed similarities exceeding 90%, as determined by any of the three methodologies - substructural similarity computation using the Tanimoto coefficient, scaffold similarity computation involving the Tanimoto scaffold similarity coefficient, or measuring SMILES string similitude through Levenshtein distance - are considered to have significant structural likeness. A difference in bioactivity of at least one order of magnitude (i.e., 10× fold disparity) in reported $K_i$ values is adopted as the threshold for defining activity cliff pairs. Similar to prior work, molecules were clustered based on substructure similarity using extended connectivity fingerprints (ECFPs). Each cluster was then split into a training set (80%) and a testing set (20%) using stratified random sampling based on the activity cliff label. Besides, we apply the scaffold splitting to construct the training (80%), validation (10%), and test sets (10%) to evaluate the generalization of MaskMol. There is only one cliff molecule in the test set of ABL11 under scaffold splitting, therefore, we do not report the $RMSE_{Cliff}$ in **Figure 2a.**

### 4.6 Compound Potency Prediction Dataset

Following Janela et al.[82], activity classes are extracted from the ChEMBL database (version 30)[83]. The selection criteria involved choosing bioactive compounds that have the highest level of confidence in direct interactions with a human target protein (confidence score 9) and a specified potency value ($IC_{50}$). The potency values were transformed into a negative logarithmic scale ($pIC_{50}$). Here, we apply the scaffold splitting to construct the training (80%), validation (10%), and test (10%) sets.

**Acknowledgements**

The work was supported by National Natural Science Foundation of China (Grant Nos. 62450002, 62202153, 62272151, 62372159, 62302156, 61972138, 62106073, 62122025, 62102140 and U22A2037), Hunan Provincial Natural Science Foundation of China (Grant

Nos. 2024JJ4015, 2023JJ40180, 2022JJ20016 and 2021JJ10020), The Science and Technology Innovation Program of Hunan Province (Grant Nos. 2022RC1100, 2022RC1099), Postgraduate Scientific Research Innovation Project of Hunan Province (Grant No. CX20220380), the project of Hunan Provincial Key Laboratory of Anti-Resistance Micro-bial Drugs (No:2023TP1013).

**Author contributions**

X.Z. and C.D. conceived the study. Z.C. and H.X. implemented the pipeline, constructed the databases, developed the codes, and performed all experiments. Z.C., H.X., P.M., J.L, X.J, X.Y, B.S., L.Z., X.F., Y.D. and X.Z. performed data analyses. Z.C., C.D., H.X., and X.Z. discussed and interpreted all results. Z.C., C.D., H.X., and X.Z. wrote and critically revised the manuscript. All authors contributed to the manuscript.

**Competing interests**

The authors declare no competing financial interest.

**References**

[1] N. Fleming, *Nature* **2018**, 557, 7706 S55.
[2] X. Zeng, F. Wang, Y. Luo, S.-g. Kang, J. Tang, F. C. Lightstone, E. F. Fang, W. Cornell, R. Nussinov, F. Cheng, *Cell Reports Medicine* **2022**.
[3] J.-P. Vert, *Nature Biotechnology* **2023**, 1–2.
[4] Y. Diao, D. Liu, H. Ge, R. Zhang, K. Jiang, R. Bao, X. Zhu, H. Bi, W. Liao, Z. Chen, et al., *Nature Communications* **2023**, 14, 1 4552.
[5] D. Flam-Shepherd, K. Zhu, A. Aspuru-Guzik, *Nature Communications* **2022**, 13, 1 3293.
[6] O. Mahmood, E. Mansimov, R. Bonneau, K. Cho, *Nature communications* **2021**, 12, 1 3156.
[7] X. Yang, L. Fu, Y. Deng, Y. Liu, D. Cao, X. Zeng **2023**.
[8] W. Jin, R. Barzilay, T. Jaakkola, *In International conference on machine learning*. PMLR, **2018**

2323–2332.

[9] W. Jin, R. Barzilay, T. Jaakkola, *In International conference on machine learning*. PMLR, **2020** 4839–4848.

[10] J. Xia, C. Zhao, B. Hu, Z. Gao, C. Tan, Y. Liu, S. Li, S. Z. Li, *In The Eleventh International Conference on Learning Representations*. **2022**.

[11] D. Xue, H. Zhang, D. Xiao, Y. Gong, G. Chuai, Y. Sun, H. Tian, H. Wu, Y. Li, Q. Liu, *bioRxiv* **2020**, 2020–12.

[12] Z. Zhang, Q. Liu, H. Wang, C. Lu, C.-K. Lee, *Advances in Neural Information Processing Systems* **2021**, 34 15870.

[13] Y. You, T. Chen, Y. Sui, T. Chen, Z. Wang, Y. Shen, *Advances in neural information processing systems* **2020**, 33 5812.

[14] S. Luo, T. Chen, Y. Xu, S. Zheng, T.-Y. Liu, L. Wang, D. He, *arXiv preprint arXiv:2210.01765* **2022**.

[15] Z. Guo, P. Sharma, A. Martinez, L. Du, R. Abraham, *arXiv preprint arXiv:2109.08830* **2021**.

[16] H. Li, R. Zhang, Y. Min, D. Ma, D. Zhao, J. Zeng, *Nature Communications* **2023**, 14, 1 7568.

[17] X. Zeng, H. Xiang, L. Yu, J. Wang, K. Li, R. Nussinov, F. Cheng, *Nature Machine Intelligence* **2022**, 1–13.

[18] J. B. Hendrickson, *Science* **1991**, 252, 5009 1189.

[19] D. Stumpfe, J. Bajorath, *Journal of medicinal chemistry* **2012**, 55, 7 2932.

[20] A. J. Wedlake, M. Folia, S. Piechota, T. E. Allen, J. M. Goodman, S. Gutsell, P. J. Russell, *Chemical Research in Toxicology* **2019**, 33, 2 388.

[21] D. van Tilborg, A. Alenicheva, F. Grisoni, *Journal of Chemical Information and Modeling* **2022**,62,

[22] J. Deng, Z. Yang, H. Wang, I. Ojima, D. Samaras, F. Wang, *arXiv preprint arXiv:2209.13492* **2022**.

[23] J. Xia, L. Zhang, X. Zhu, S. Z. Li, *arXiv preprint arXiv:2306.17702* **2023**.

[24] T. N. Kipf, M. Welling, *arXiv preprint arXiv:1609.02907* **2016**.

[25] P. Veličković, G. Cucurull, A. Casanova, A. Romero, P. Lio, Y. Bengio, *arXiv preprint arXiv:1710.10903* **2017**.

[26] J. Gilmer, S. S. Schoenholz, P. F. Riley, O. Vinyals, G. E. Dahl, *In International conference on ma-chine learning*. PMLR, **2017** 1263–1272.

[27] Q. Li, Z. Han, X.-M. Wu, *In Proceedings of the AAAI conference on artificial intelligence*, volume 32. **2018** .

[28] K. He, X. Zhang, S. Ren, J. Sun, *In Proceedings of the IEEE conference on computer vision and pattern recognition*. **2016** 770–778.

[29] D. Rogers, M. Hahn, *Journal of chemical information and modeling* **2010**, 50, 5 742.

[30] J. Devlin, M.-W. Chang, K. Lee, K. Toutanova, *arXiv preprint arXiv:1810.04805* **2018**.

[31] K. He, X. Chen, S. Xie, Y. Li, P. Dollár, R. Girshick, *In Proceedings of the IEEE/CVF Conference on Computer Vision and Pattern Recognition*. **2022** 16000–16009.

[32] A. Dosovitskiy, L. Beyer, A. Kolesnikov, D. Weissenborn, X. Zhai, T. Unterthiner, M. Dehghani, M. Minderer, G. Heigold, S. Gelly, et al., *arXiv preprint arXiv:2010.11929* **2020**.

[33] W. Kim, B. Son, I. Kim, *In International Conference on Machine Learning*. PMLR, **2021** 5583–5594.

[34] T. Brown, B. Mann, N. Ryder, M. Subbiah, J. D. Kaplan, P. Dhariwal, A. Neelakantan, P. Shyam,

G. Sastry, A. Askell, et al., *Advances in neural information processing systems* **2020**, 33 1877.

[35] A. Radford, K. Narasimhan, T. Salimans, I. Sutskever, et al., *OpenAI blog* **2018**.

[36] Y. Hu, J. Bajorath, *Journal of chemical information and modeling* **2012**, 52, 7 1806.

[37] D. Stumpfe, H. Hu, J. Bajorath, *Journal of Computer-Aided Molecular Design* **2020**, 34 929.

[38] S. Chithrananda, G. Grand, B. Ramsundar, *arXiv preprint arXiv:2010.09885* **2020**.

[39] Y. Rong, Y. Bian, T. Xu, W. Xie, Y. Wei, W. Huang, J. Huang, *Advances in Neural Information Processing Systems* **2020**, 33 12559.

[40] Y. Wang, J. Wang, Z. Cao, A. Barati Farimani, *Nature Machine Intelligence* **2022**, 4, 3 279.

[41] X. Fang, L. Liu, J. Lei, D. He, S. Zhang, J. Zhou, F. Wang, H. Wu, H. Wang, *Nature Machine Intelligence* **2022**, 4, 2 127.

[42] W. Hu, B. Liu, J. Gomes, M. Zitnik, P. Liang, V. Pande, J. Leskovec, *arXiv preprint arXiv:1905.12265* **2019**.

[43] H. Stärk, D. Beaini, G. Corso, P. Tossou, C. Dallago, S. Günnemann, P. Liò, *In International Conference on Machine Learning.* PMLR, **2022** 20479–20502.

[44] S. Liu, H. Wang, W. Liu, J. Lasenby, H. Guo, J. Tang, *arXiv preprint arXiv:2110.07728* **2021**.

[45] F. Wu, H. Qin, W. Gao, S. Li, C. W. Coley, S. Z. Li, X. Zhan, J. Xu, *arXiv preprint arXiv:2304.03906* **2023**.

[46] N. Cristianini, J. Shawe-Taylor, *An introduction to support vector machines and other kernel-based learning methods,* Cambridge university press, **2000**.

[47] L. Breiman, Machine learning **1996**, 24 123.

[48] E. Fix, J. L. Hodges, *International Statistical Review/Revue Internationale de Statistique* **1989**, 57,3 238.

[49] Y. LeCun, Y. Bengio, G. Hinton, *nature* **2015**, 521, 7553 436.

[50] J. H. Friedman, *Annals of statistics* **2001**, 1189–1232.

[51] S. Kullback, R. A. Leibler, *The annals of mathematical statistics* **1951**, 22, 1 79.

[52] D.-H. Lee, et al., *In Workshop on challenges in representation learning*, ICML, volume 3.Atlanta,**2013** 896.

[53] W. Torng, R. B. Altman, *Journal of chemical information and modeling* **2019**, 59, 10 4131.

[54] M. Sakai, K. Nagayasu, N. Shibui, C. Andoh, K. Takayama, H. Shirakawa, S. Kaneko, *Scientific reports* **2021**, 11, 1 525.

[55] P. Li, J. Wang, Y. Qiao, H. Chen, Y. Yu, X. Yao, P. Gao, G. Xie, S. Song, *Briefings in Bioinformatics* **2021**, 22, 6 bbab109.

[56] M. P. Fay, M. A. Proschan, *Statistics surveys* **2010**, 4 1.

[57] R. R. Selvaraju, M. Cogswell, A. Das, R. Vedantam, D. Parikh, D. Batra, *In Proceedings of the IEEE international conference on computer vision.* **2017** 618–626.

[58] D. Luo, W. Cheng, D. Xu, W. Yu, B. Zong, H. Chen, X. Zhang, *Advances in neural information processing systems* **2020**, 33 19620.

[59] Z. Wu, J. Wang, H. Du, D. Jiang, Y. Kang, D. Li, P. Pan, Y. Deng, D. Cao, C.-Y. Hsieh, et al., *Nature Communications* **2023**, 14, 1 2585.

[60] Z. Wu, D. Jiang, J. Wang, C.-Y. Hsieh, D. Cao, T. Hou, *Journal of Medicinal Chemistry* **2021**, 64,10 6924.

[61] C. Xu, F. Cheng, L. Chen, Z. Du, W. Li, G. Liu, P. W. Lee, Y. Tang, *Journal of chemical information and modeling* **2012**, 52, 11 2840.

[62] P. G. Polishchuk, V. E. Kuz'min, A. G. Artemenko, E. N. Muratov, *Molecular Informatics* **2013**,

32, 9-10 843.

[63] S. Peng, P. Hu, Y.-T. Xiao, W. Lu, D. Guo, S. Hu, J. Xie, M. Wang, W. Yu, J. Yang, et al., *Clinical Cancer Research* **2022**, 28, 3 552.

[64] T. Liu, Y. Lin, X. Wen, R. N. Jorissen, M. K. Gilson, *Nucleic acids research* **2007**, 35, suppl 1 D198.

[65] K. Nakao, A. Murase, H. Ohshiro, T. Okumura, K. Taniguchi, Y. Murata, M. Masuda, T. Kato, Y. Okumura, J. Takada, *Journal of Pharmacology and Experimental Therapeutics* **2007**, 322, 2 686.

[66] J. He, X. Lin, F. Meng, Y. Zhao, W. Wang, Y. Zhang, X. Chai, Y. Zhang, W. Yu, J. Yang, et al., *Molecules* **2022**, 27, 4 1209.

[67] A. Murase, T. Okumura, A. Sakakibara, H. Tonai-Kachi, K. Nakao, J. Takada, *European journal of pharmacology* **2008**, 580, 1-2 116.

[68] M. Blouin, Y. Han, J. Burch, J. Farand, C. Mellon, M. Gaudreault, M. Wrona, J.-F. L'evesque, D. Denis, M.-C. Mathieu, et al., *Journal of medicinal chemistry* **2010**, 53, 5 2227.

[69] G. Caselli, A. Bonazzi, M. Lanza, F. Ferrari, D. Maggioni, C. Ferioli, R. Giambelli, E. Comi, S. Zerbi, M. Perrella, et al., *Arthritis Research & Therapy* **2018**, 20 1.

[70] T. Kotani, H. Takano, T. Yoshida, R. Hamasaki, G. Kohanbash, K. Takeda, H. Okada, *Cancer Research* **2020**, 80, 16 Supplement 4443.

[71] D. I. Albu, Z. Wang, K.-C. Huang, J. Wu, N. Twine, S. Leacu, C. Ingersoll, L. Parent, W. Lee, D. Liu, et al., *Oncoimmunology* **2017**, 6, 8 e1338239.

[72] Y. Jin, Q. Liu, P. Chen, S. Zhao, W. Jiang, F. Wang, P. Li, Y. Zhang, W. Lu, T. P. Zhong, et al., *Cell Discovery* **2022**, 8, 1 24.

[73] D. Das, D. Qiao, Z. Liu, L. Xie, Y. Li, J. Wang, J. Jia, Y. Cao, J. Hong, *ACS Medicinal Chemistry Letters* **2023**, 14, 6 727.

[74] N. Salim, J. Holliday, P. Willett, *Journal of chemical information and computer sciences* **2003**, 43,2 435.

[75] M. J. McGregor, S. M. Muskal, *Journal of chemical information and computer sciences* **1999**, 39, 3569.

[76] A. M. Schreyer, T. Blundell, *Journal of cheminformatics* **2012**, 4 1.

[77] J. Degen, C. Wegscheid-Gerlach, A. Zaliani, M. Rarey, ChemMedChem: *Chemistry Enabling Drug Discovery* **2008**, 3, 10 1503.

[78] S. Kim, J. Chen, T. Cheng, A. Gindulyte, J. He, S. He, Q. Li, B. A. Shoemaker, P. A. Thiessen, B. Yu, et al., *Nucleic acids research* **2019**, 47, D1 D1102.

[79] T. Janela, J. Bajorath, *Nature Machine Intelligence* **2022**, 4, 12 1246.

[80] A. P. Bento, A. Gaulton, A. Hersey, L. J. Bellis, J. Chambers, M. Davies, F. A. Krüger, Y. Light, L. Mak, S. McGlinchey, et al., *Nucleic acids research* **2014**, 42, D1 D1083.

[81] C. Chen, W. Ye, Y. Zuo, C. Zheng, S. P. Ong, *Chemistry of Materials* **2019**, 31, 9 3564.

[82] A. Krizhevsky, I. Sutskever, G. E. Hinton, *Advances in neural information processing systems* **2012**,25.

[83] S. Hochreiter, J. Schmidhuber, *Neural computation* **1997**, 9, 8 1735.

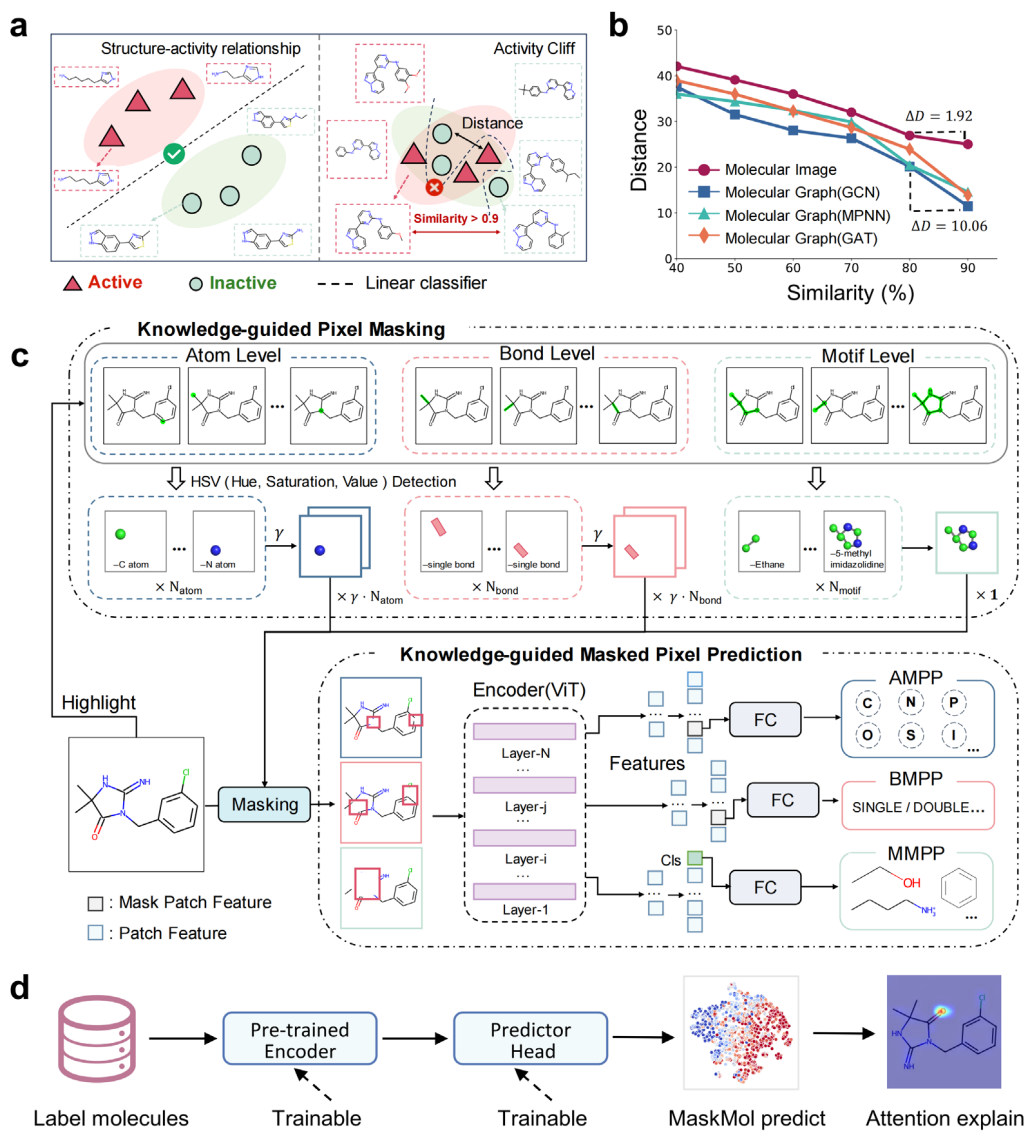

**Figure 1.** Overview of the MaskMol framework. **(a)** Examples of SAR *(Left)* and activity cliffs *(Right)* in feature space. Highly active molecules are laid in red boxes and low-active molecules in green boxes. **(b)** Comparison of graph and image in feature space. The similarity is calculated using the Tanimoto coefficient on molecular ECFP fingerprint pairs[29]. The distance is defined as the average Euclidean distance in the 2D space of the 1000 molecule pairs after the encoder extracts features. Among them, the molecular image uses ResNet18[28], and the molecular Graph adopts GCN, MPNN, and GAT as the encoder. **(c)** The framework of our MaskMol pipelines. The framework comprises two major components, knowledge-guided pixel masking and knowledge-guided masked pixel prediction. RDKit generates molecular highlight images and HSV color detection is used to obtain molecular knowledge-guided masked images. In the pre-training stage, self-supervised learning tasks of masking are introduced to capture knowledge and structure information hidden inside the molecular image. **(d)** The finetuning of MaskMol on downstream benchmarks (such as activity cliff estimation and compound potency prediction), where the parameters of the pre-trained encoder and predictor head are trainable.

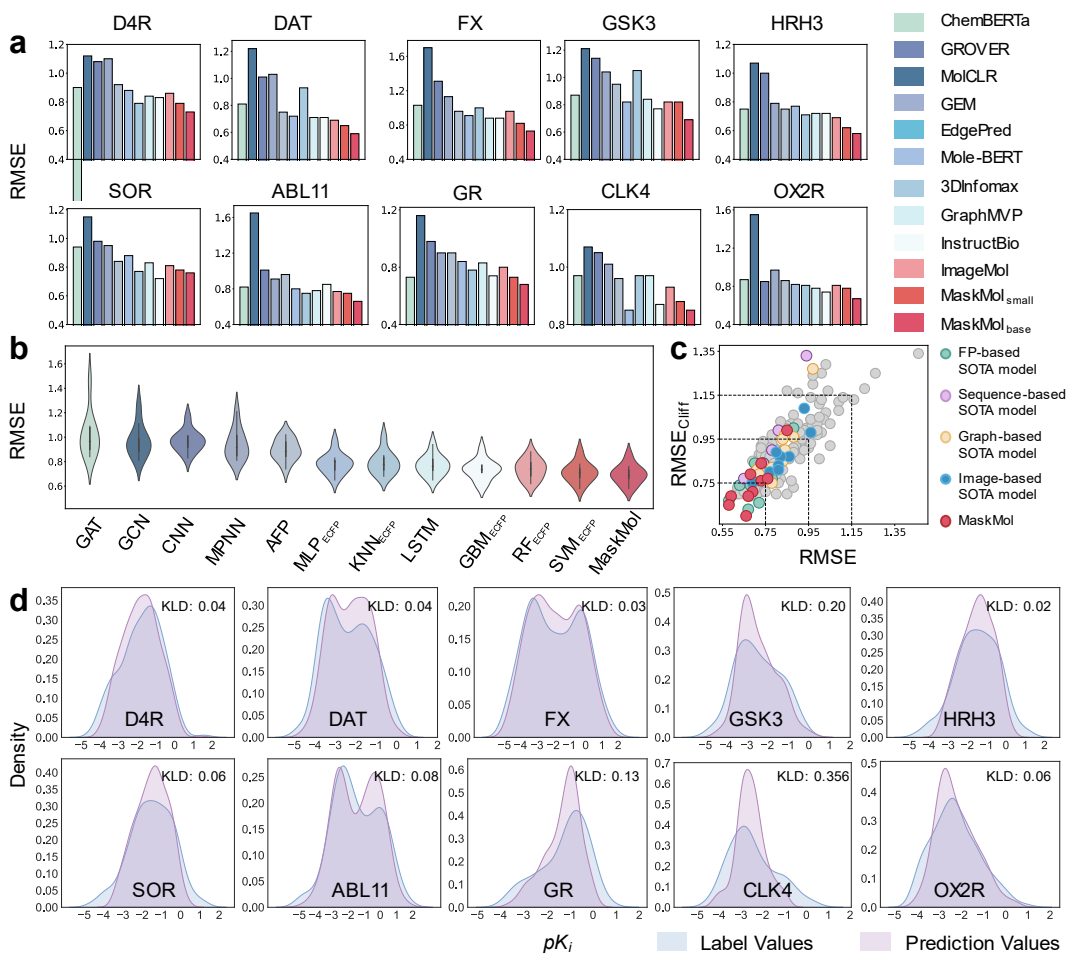

**Figure 2.** Performance of the MaskMol framework on activity cliff estimation (ACE). **(a)** Comparison of the error on all compounds (denoted as RMSE) on 10 ACE datasets from MoleculeACE with different representation pre-training methods. **(b)** The violin plot displays RMSE comparison with traditional machine learning methods. In a violin plot, a value distribution is represented by its maximum value (upper thin line), upper quartile (upper thick line), median value (white dot), lower quartile (lower thick line), and minimum value (lower thin line). **(c)** Comparison between the error on activity cliffs compounds (denoted as $RMSE_{Cliff}$ and RMSE for all methods. Green, purple, yellow, and blue represent the SOTA methods based on fingerprint ($SVM_{ECFP}$), sequence (LSTM), graph (GraphMVP), and image (ImageMol), respectively. **(d)** Evaluation the gap between label values and prediction values on ACE, in which the data distribution is determined by a kernel density estimation. The kullback-leibler divergence (KLD) is to measure the difference between two probability distributions.

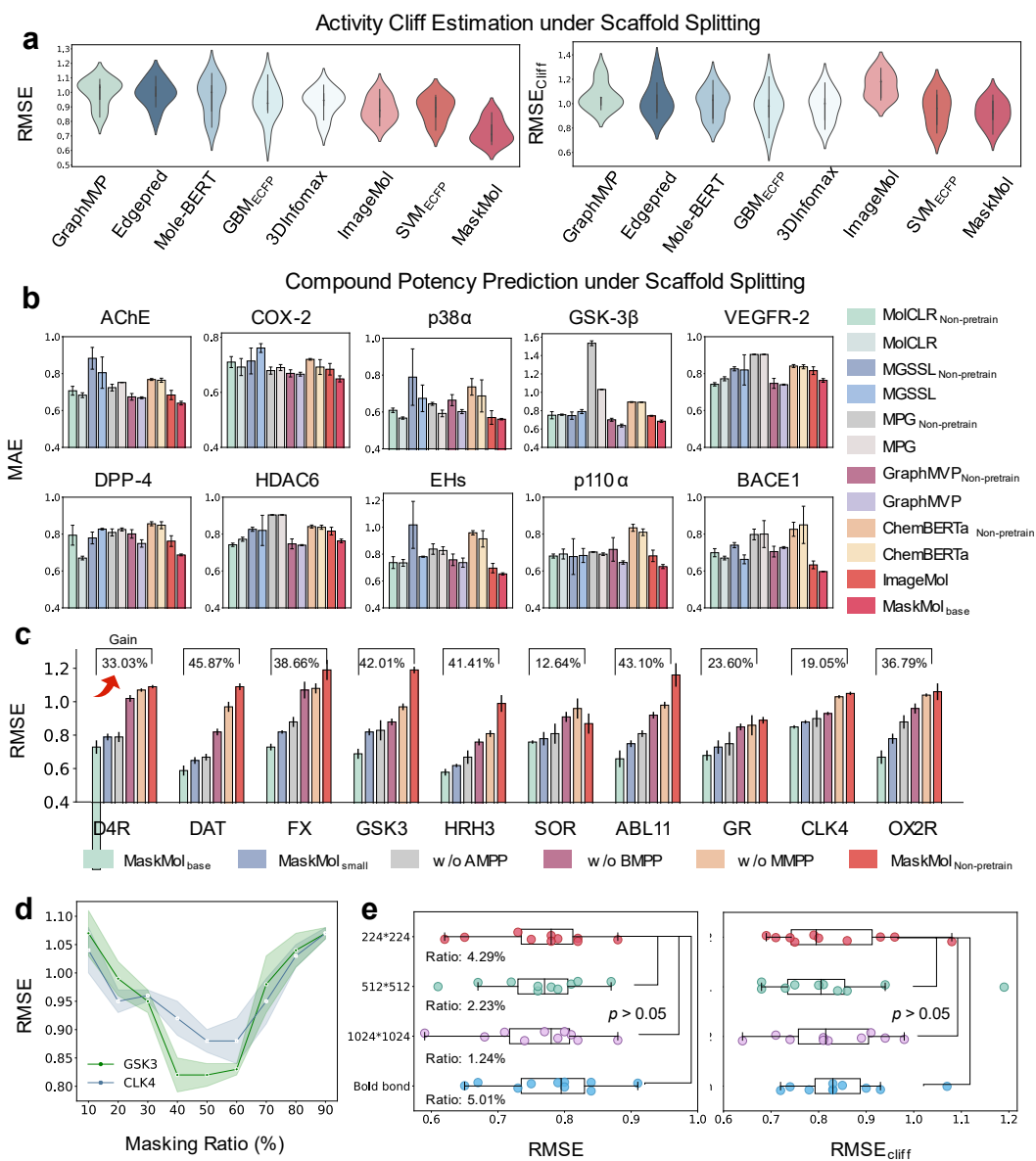

**Figure 3.** Performance of the MaskMol in the scaffold splitting scenarios and ablation study. **(a-b)** The performance of MaskMol and baseline methods on **(a)** ACE task and **(b)** compound potency prediction (CPP) task, measured in terms of RMSE and MAE, respectively. All the prediction results were reported based on three independent runs on three random seeded(seed=0,1,2). Data are presented as mean ± standard deviation (SD). **(c)** The ablation study of the pretext task in the MoleculeACE setting. "w/o AMPP" denotes removing the AMPP pretext task component during pre-training, "w/o BMPP" and "w/o MMPP" in the same way. "Gain" indicates the improvement of the MaskMol$_{base}$ compared to the MaskMol$_{Non-Pretrain}$. **(d)** Ablation study on the masking ratio used in the pre-training stage. The x-axis represents the masking ratio in the AMPP and BMPP tasks. The error bands stand for the standard deviations. **(e)** Experiments on different image sizes and useful pixel ratios in ACE datasets. The ratio is the molecule pixels divided by the whole image pixels on 12,590 images. RDKit is used to bold the chemical bonds to generate an image with a higher ratio of useful pixels. Statistical analysis is performed with the Mann-Whitney U test.

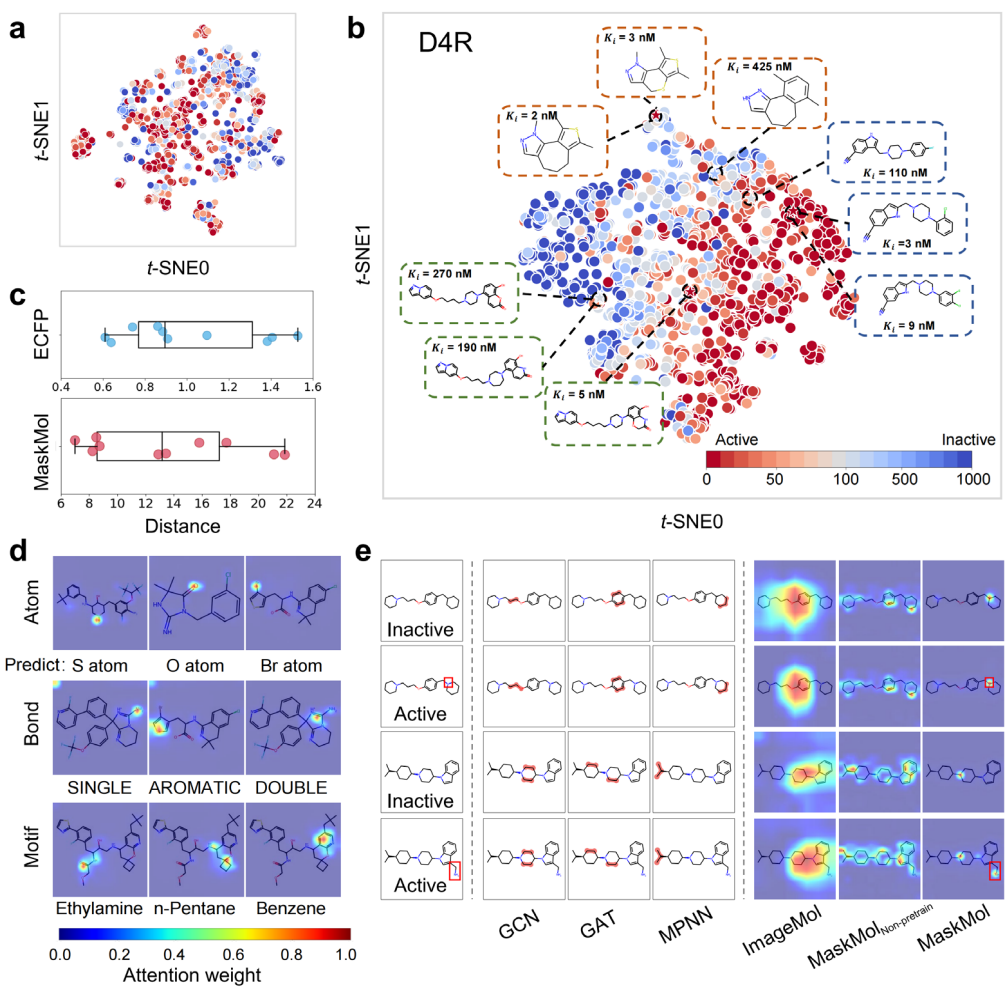

**Figure 4.** Feature distribution and Attention interpretation of MaskMol. **(a-b)** Visualization of molecular ECFP fingerprint **(a)** and molecular representations derived from MaskMol **(b)** via t-SNE on D4R dataset from ACE task, respectively. Each point is colored according to its corresponding $K_i$ value. The larger the $K_i$ value, the cooler the color, and the smaller the corresponding $K_i$ value, the warmer the color. Each pair of structurally similar but bioactivity-distinct molecules are laid in boxes with the same color. **(c)** Comparison of ECFP fingerprint and MaskMol in quantifying the overall relative distance of active cliff pairs in the latent features space. **(d)** Examples of MaskMol's heatmaps of three knowledge levels highlighted by Grad-CAM. A hotter color area indicates higher MaskMol attention. Given a masked image with different knowledge levels, MaskMol gives the masked knowledge type and focuses on the correct mask area. **(e)** Visualization of the explanatory structure found by different DL methods on activity cliffs. Image-based methods use Grad-CAM to visualize attention, while graph-based approaches employ PGExplainer to mine essential structures.

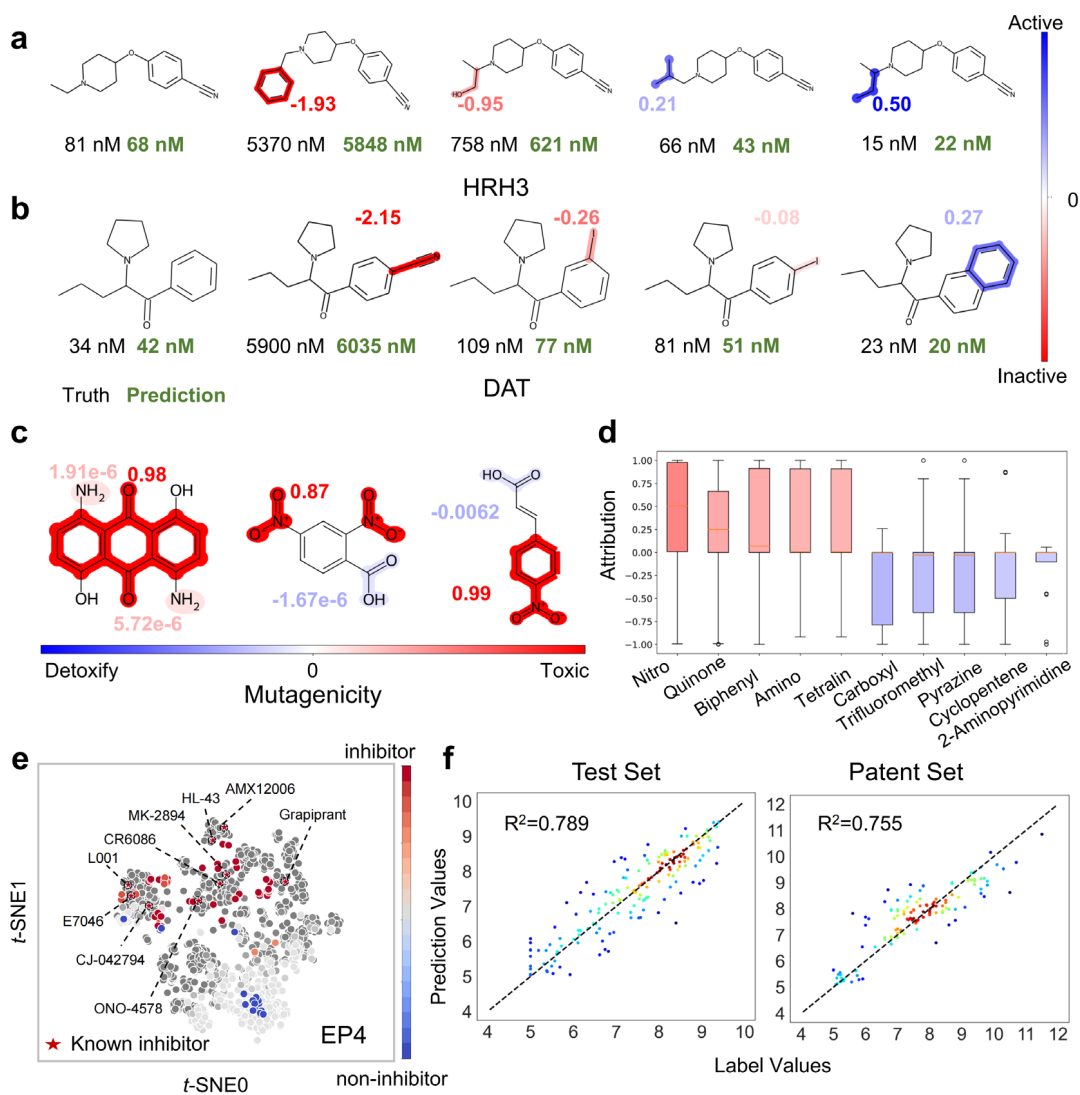

**Figure 5.** Chemistry-intuitive explanation of MaskMol and virtual screening on EP4 target. **(a-b)** and **(c-d)**, are MaskMol's interpretations of biological activity and toxicity, respectively. **(a-b)** The attribution visualization of four compounds of HRH3 and DAT receptors, respectively. Black represents the true value and green represents the prediction value. **(c)** The attribution visualization of three compounds of mutagenicity. **(d)** The attribution of the different functional groups in the whole dataset. Only the functional groups that appear more than twenty times in the dataset are included. Blue hues indicate negative attributions below zero, signifying the substructure's inclination towards the property, while red hues represent positive attributions above zero, implying the substructure's unfavorable impact on the property. The mutagenicity model achieved high ROC-AUC scores of 0.90. **(e)** Visualization of molecular representations of molecules from the EP4 derived from MaskMol. Grey dots represent molecules used to train MaskMol, and dark grey dots represent inhibitors. Red and blue dots represent inhibitors and non-inhibitors in the patent set, respectively. The dashed circles indicate that the molecules had been previously identified as inhibitors of EP4 in previous studies. **(f)** The gap between prediction values and label values of test set (*Left*) and patent set (*Right*). The black dotted line represents y=x, and the closer to this line, the warmer the point color.

33